\documentclass[journal]{IEEEtran}
\usepackage[pdftex]{graphicx}
\usepackage[utf8]{inputenc}
\usepackage{cite}
\usepackage{enumitem}
\usepackage{url}
\usepackage{balance}
\begin{document} 
\title{Exploring Dynamic Difficulty\\Adjustment in Videogames}
\author{Gabriel~K.~Sepulveda, Felipe~Besoain, and Nicolas~A.~Barriga
\thanks{G.K. Sepulveda, F. Besoain, and N.A. Barriga are with the Escuela de
Ingeniería en Desarrollo de Videojuegos y Realidad Virtual, Facultad de
Ingeniería, at Universidad de
Talca. Campus Talca, Chile. (e-mail: \mbox{gsepulveda17@alumnos.utalca.cl},
fbesoain@utalca.cl, nbarriga@utalca.cl).}
\thanks{Corresponding Author: N.A. Barriga. (e-mail: nbarriga@utalca.cl)}}

\maketitle 
\begin{abstract} 
%problem
Videogames are nowadays one of the biggest entertainment industries in the world. 
Being part of this industry means competing against lots of other companies and 
developers, thus, making fanbases of vital importance. They are a group of
clients that
constantly support your company because your video games are fun. Videogames 
are most entertaining when the difficulty level is a good match for the
player's skill, increasing
the player engagement. However, not all players are equally proficient, so some kind 
of difficulty selection is required.
%solution
In this paper, we will present 
%the problem of
 Dynamic Difficulty
Adjustment~(DDA), a recently arising
research topic, which aims to develop an automated difficulty selection
mechanism that keeps the player engaged and properly challenged, neither bored
nor overwhelmed. We will present some recent research addressing this issue, as
well as an overview of how to implement it.
%impact
Satisfactorily solving the DDA problem directly affects the player's experience
when 
playing the game, making it of high interest to any game developer, from
independent ones, to 100 billion dollar businesses, because of
the potential impacts in player retention and monetization.

\end{abstract}


\section{Introduction} 

%presentacion
\IEEEPARstart{W}{hen} someone is playing a videogame, his goal is to have fun. Game developers
must infuse the games with this fun. The fun factor is composed of four axes:
fantasy, curiosity, player control, and challenge. If kept in balance the player
will stay entertained~\cite{Malone1981Toward}. The challenge axis is the most
difficult to control because to do so, the game difficulty and player skill must
match.

%problema 
Setting a single difficulty level fit for every player is not possible. A
solution is to allow the player to choose a difficulty level. This method has
some drawbacks, such as having a limited set of difficulty levels, creating gaps
where
players can fall and having difficulty progression that mismatch player learning
curves. Also, by making the player aware of the change in difficulty the game
experience is affected. Over the last decade, there have been multiple
publications related to
the improvement of this issue, using Dynamic Difficulty Adjustment (DDA)~\cite{Mohammad2018Dynamic}. 
As a result, there are various algorithms the
developer can use to implement DDA, and choosing the most appropriate one can be
a difficult task.
 The lack of experience makes choosing
and correctly applying the algorithm harder, leading to a poor implementation
causing conflicts in the game.

In the following section we will briefly describe the Dynamic Difficulty
Adjustment problem. Section~\ref{sec:skill} will introduce the readers to the
assessment of player's skill levels. Section~\ref{sec:impl} gives an overall explanation
of how a DDA system works, while section~\ref{sec:mod} reviews the
implementations of different approaches to DDA. Finally, we close with a summary
and some pointers for future work.

\section{Background} \label{sec:back}

When doing an activity that is neither boring nor frustrating, the person
becomes engrossed in said activity, being able to perform longer and keep
focused on the task. This state of mind is called the flow channel (flow)
\cite{Csikszentmihalyi2009Flow} and is present in all fields. This concept was
 later on introduced in the videogames area by Koster~\cite{koster2013theory}.

\begin{figure}[!b]
\includegraphics[width=\columnwidth]{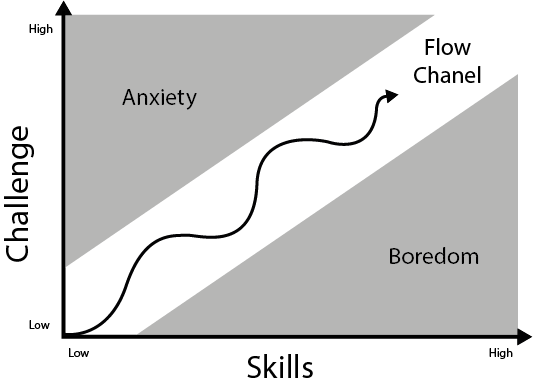}
\caption{Flow Channel}
\label{fig:flow}
 \end{figure}

In figure~\ref{fig:flow}, it's possible to note that when the difficulty of the
game is
higher than the players skills the activity becomes frustrating pushing the
player into a state of anxiety. In contrast when the player skills are higher
than the difficulty, the game is too easy, pushing the player into a state of
boredom. When neither of those happen, the user is faced by a challenge whose
difficulty level matches the player's skill, enabling him to enter the flow.
Providing a series of challenges allows the player to stay in the flow for 
longer periods of time.

 It is important to note that taking breaks between the
challenges will prevent overwhelming the player. By alternating a series of
constant challenges with break times it's possible to create a game that
enthralls the player and keeps him playing.

Hunicke proposes the creation of a system with techniques for stochastic
 inventory management that periodically examines the player progress and 
dynamically adjust the difficulty of the game challenges to adjust the player's 
overall experience~\cite{hunicke2004ai}.
Later on, he observes that a DDA system can improve the player experience
 without the need for any sophisticated AI~\cite{hunicke2005case}.
    
DDA is an AI-based system that allows the change of attributes and behaviors
within the game in runtime. DDA measure the player performance and change the
game difficulty to match the player skills. As a result it creates the
challenge used to guide the player into the flow zone.
A good DDA must be able
to track the player skill level and adapt to it. Changes in difficulty must
follow the player learning curve and go unnoticed by the
player~\cite{Mohammad2018Dynamic}.

A DDA system that fulfills these requirements can increase the player 
confidence in his chances to beat the game when presented with a hard 
challenge~\cite{Constant2019Dynamic} and the core engagement of the player, 
resulting in an increased gameplay duration~\cite{xue2017dynamic}.

\section{Assessing Player Skills}\label{sec:skill}
 
The first step needed to implement a DDA system is an evaluation of the player's
performance. Through a set of predefined variables it's possible to asses if the
difficulty is fit for the player. This is done by comparing the current in-game
value of this variables with their expected value. 

 \subsection{Variables}

In order to choose the variables used to evaluate the player performance it is
important to have a clear notion of what is considered as a failure, a success
or a skill that the player needs to develop within the game. Usually, they'll correspond to 
game rules and win or loss conditions~\cite{Hendrix2018Implementing}.

A good example is given in a study using Pac-Man as a test-bed. By measuring the
number of hits on the maze walls, the number of keys pressed and the number of
direction switches, it is possible to have an idea of what the player skills
are~\cite{Ebrahimi2014Dynamic}. Also the lives lost and pills collected versus
time are good indicators of failure and success rate.

Another example is given by the analysis performed to the Multiplayer Online
Battle Arena (MOBA) game Defense of the Ancient 
(DotA)\footnote{\url{https://en.wikipedia.org//wiki/Defense_of_the_Ancients}}
 In this case the level
reached versus time, and towers destroyed versus time, are used to measure success,
and deaths versus time, used for failure~\cite{Silva2015Dynamic}. Additionally,
we can count the number of minions killed, heroes killed, items completed and
missed abilities, among others, to check the player skill level.

As in the previous examples, all games can include variables that indicate
the
current state of the player and give information about his performance, success and failure 
ratio, and his learning process. The variables selected will depend on the specific game in
 which the DDA system is implemented. 

Note that a high number of
variables will result in a more precise difficulty adjustment but will also
consume more memory, in the other hand a low number of variables will result in
poor adjustment. The amount of variables chosen will depend on the complexity
of the game but for most of the studies a number of three to five variables is
enough.

\subsection{Data Collection} 

For the DDA system to keep track of the chosen variables, the game has to
overwrite their value. The tracking can be event-triggered or
permanent.

Event-triggered tracking is for variables that change when fulfilling a
condition or performing an action. In this case, the method in charge of
triggering the event can call the DDA system class and change the variable
value.

An example of this are the variables used in the previously mentioned Pac-Man
research~\cite{Ebrahimi2014Dynamic}. The
number of hits on maze walls, number of keys pressed or number of direction
changes must be actualized just when the corresponding action happens.

On the other hand, permanent tracking is applied to variables that are always in
 game. The game has to call the DDA system class and update the values of
the variables in the game loop. Setting a minimum time between updates is
recommended to reduce processing.

Examples of variables that need permanent tracking are player health, player
gold, and game progress.
    
\subsection{Reference Point} 

With the collected data, it’s possible to assess the player skills using the 
performance of a player that matches the game difficulty as a reference point.
The chances of finding a player that perfectly matches the game difficulty are 
zero. Thus, we need to find a method that can provide the 
reference point.

Setting the reference point based on the beliefs of what it should be is the 
worst way to do so. Even with years of experience, it's unlikely to guess 
the correct value.

One option is to use an AI agent capable of playing the game. The data of the 
AI play-through will be then used as a reference point. In order to get reliable
 information the agent must play the game many times, a bigger number of
 iterations gives more precise information. An agent that plays the game perfectly 
isn’t useful as it has to imitate a real player, to that end an agent that can
play at a medium level is required instead.

However, the best option would be to have real player data. Using the same
system
to get data the game can be tested with real players. Using a survey after 
the testing session, the data of players that had fun while playing can be used
 as a sample. 
    
\subsection{Data Analysis}

Having set the varables to analyze, the reference point for each, and being
able to save the data from the player play-through, the system has the
essentials needed
for the evaluation of player performance. These evaluations should be made
through the use of methods that return a success, failure or skill ratio. Just
subtracting the number of player deaths from the expected player deaths returns
a non-representative number in most cases. 
The evaluation must also happen
constantly in the game, for the difficulty to change at the right times. To
this end, the evaluation of the player performance must happen once each X
number of ticks. A low X value makes the game more adaptable, but increase the
process cost of the game and might not be needed, selecting the right X value will
depend on the game genre.

Evaluation of player performance can be done by comparing the player's current
stats with the ideal or reference ones versus a delta 
time~\cite{Ebrahimi2014Dynamic} (equation~\ref{eq:diff}).

\begin{equation}\label{eq:diff}
 Difficulty = (N - Z)/D;
\end{equation}
\begin{equation}
Ease = 1 - Difficulty;
\end{equation}

Let’s take, for example, a situation where the player performs an action
 $N$ times in a $D$ period of time or ticks, with the ideal
or reference value being $Z$. Using a value set that abides by $0 \leq (N-Z) \leq
D$ will return a normalized value, allowing an easier interpretation of the results.

As the enjoyment of a game is greater when the game is equally
hard and easy, the point where these two values merge is the desired game state.
A return value close to 0.5 fulfills this condition. The chances of getting the desired 
value are low, that’s why leaving a proper margin is recommended. 
The margin must be defined by the game developer based on the situation. Leaving a
 0.1 error margin would leave us with a range between [0.4, 0.6], where the game 
difficulty is acceptable.

This method also allows calculating the player global proficiency by calculating
the average of all variables or a weighting of them.
Consider that it is convenient for the global proficiency, or player global
performance, to be a normalized value. To this end, the sum
of all weights must also be normalized.

Another method is to assess the player current performance (Cp) versus the
expected player performance at that time (Ep[t]), which would be our reference
point.

\begin{equation}
 Performance = Cp/Ep[t];
\end{equation}

In this manner, a value next to 1 means that the challenge is appropriate for
the player. Same as in the previous method a error margin defined by the game developer
 is needed as getting the exact value is unlikely.
For example, setting the error margin of 0.2 give us a range [0.8, 1.2]. This
mean that values
lower than 0.8 indicates the game is too hard and values greater than 1.2 means
the game is too easy.

This method also allows to create a ranking of the player global performance by
adding the results. Both methods can be modified to better suit the game
developer preferences and needs.
     
\section{Implementation}\label{sec:impl}

There are several different DDA methods proposed in the literature. We will
focus on the more straightforward ones, with the purpose of giving the reader
some insight on how to
implement the DDA system into his projects. For more 
information, a recent review of DDA research from 2009 to 2018 is 
available~\cite{Mohammad2018Dynamic}. Table~\ref{tab:app}
summarizes the approaches found in the literature.

At the start of the game, there is no data of the player
performance, and so, the developer must set the difficulty based on the
average testing results. Then, the game must quickly adapt to the player
performance. Playable tutorials are a good opportunity to get early information
on the player performance without having to make him go through any real
challenge. Afterward, the change of difficulty should happen less often and be
smaller, allowing for the growth of the player skills.

The biggest challenge while changing the difficulty of the game in real time is
avoiding the player noticing the change. To avoid this, performing the
difficulty change at times where the player is not aware is a must. Such cases
are times when the player is dead, change of scenes, features or elements the
player has not yet seen. If the change is subtle enough, elements that the
player is not currently seeing, non-perceptible changes as most of the element
specific variable values changes, or minor behavior changes can be performed
without fearing the player to notice it, as long as the changes don't happen too
often, and aren't too extreme.
There must be an upper and lower limit for how
much a variable can change, as well as a time threshold for the changes to
happen and a maximum number of changes for each update and stage. The changes to
be performed can be added to a queue as functors or callbacks to be
executed at a
given time. Each change can have a tag that identifies the change. When a change
is going to enter the queue any changes performed with the same tag will be
removed, preventing two changes to affect the same element, as this is unneeded
and can cause problems. 

\begin{table}[!t]
\renewcommand{\arraystretch}{1.3}
\caption{DDA Approaches}
\label{tab:app}
\centering
\begin{tabular}{ccc}
\hline\hline
\textbf{Author(s)} & \textbf{Year} & \textbf{Approach}\\
\hline
Hunicke and Chapman & 2004 & Hamlet System~\cite{hunicke2004ai}\\
%\hline
Spronck et al. & 2006 & Dynamic Scripting~\cite{spronck2006adaptive}\\
%\hline
Pedersen, Togelius,& 2009 & Single and\\ and Yannakakis && multi-layered perceptrons~\cite{Pedersen2009Modeling}\\
%\hline
Hagelback and Johansson & 2009 & Reinforcement Learning~\cite{Hagelback2009Measuring}\\
%\hline
&& Upper Confidence Bound \\Li et al. & 2010 & for Trees and \\&&Artificial Neural Networks~\cite{Li2010To}\\
%\hline
Ebrahimi and Akbarzadeh-T & 2014 & Self-organizing System and \\&&Artificial Neural Networks~\cite{Ebrahimi2014Dynamic}\\
%\hline
Sutoyo et al. & 2015 & Metrics~\cite{sutoyo2015dynamic}\\
%\hline
Xue et al. & 2017 & Probabilistic Methods~\cite{xue2017dynamic}\\
%\hline
Stein et al. & 2018 & EEG-triggered dynamic\\&& difficulty adjustment~\cite{stein2018eeg}\\
\hline\hline
\end{tabular}
\end{table}

Big changes, such as the size of a room, conspicuous change of
behavior with no reason in the gameplay, and others, must be executed
in load screens, while changing scenes, or in sections not seen by the player.

These can also be added to the functors or callbacks queue and be executed when needed. 
On the other hand, changes to be
performed on zones unseen to the player, but on the same scene, can be
performed immediately to prevent the player for getting to the zone with the
changes undone.

The creation of mathematical functions that tells us how much the variables
should be changed is probably the best way to perform these changes. The functor
can receive as parameter the proficiency of the player in the corresponding
field
and calculate how much each variable must change to fit the player. The
definition of the mathematical functions must be done with the data collected
from the multiple testing phases. Creating a game with preset difficulties and
make a study where players test all the difficulty levels is advisable.

Each one of the variables selected as an indicator of the player performance is 
directly affected by at least one factor in the game. For example, the death
ratio 
of the player depends on the damage dealt by the enemies, the chances to evade
 an attack, and the aggressiveness of the enemies, among others.
Linking each evaluation variable to a factor is crucial, as this factors are what is going
 to be changed in order to adapt the game difficulty to the player.
These factors can be categorized in three sections, attributes, behaviors and events.

\subsection{Attributes} 

Changing the value of attributes in the game is the first and most intuitive of
the modifications to be done, and also, the easiest of them. Even so, the
number
of attributes that can create the difficulty trait in one of these
variables can
be immense. Specific information, like the given by reports of
event-triggered parameters, can allow discerning how to properly balance the
game. By comparing the player stats with the enemy who killed him, and the time
of the engagement, it is possible to know if the player died because of the
difference in damage, speed, range, attack ratio or others. There are also
situations where the player died because his health was low before the
engagement, which means that it's not the current enemy the reason for his
death, but a previous one or the sum of them. The tracking of spikes in
permanent game values allows a better adjustment. If there was a spike drop in
health, knowing whether it was caused by a single enemy or by  a group attack,
can trigger the adjustment of the single enemy's stats, or the reduction in the
number of enemies. If there was no spike then maybe the speed at which the waves
of enemies reach the player is too fast and slowing down is required.
As shown in the previous example, the traits of the characters in the game are
not the only thing that can be changed, but also the number of enemies on the
same zone, or even subtle things, such as the auto-aim range, the time
limit for
a quick time event, the dimension of a room,
or the pace of the game.

\subsection{Behaviors} 

Traits in a game can be presented not only as
attributes but also as behaviors. Behaviors are the main component of complexity
in a game and are not exclusive to NPCs (Non-Playable Characters), in
three-in-line games the pieces have the behavior to self destruct if there are
another two equal pieces at a two tiles distance (vertical or
horizontal), in the same axis (x,y). Modern three-in-line have added new blocks
and behaviors. These behaviors cannot be changed, as they are the foundation of
the
game, but not all games have these restrictions.

In a stealth game, the watch tower can move the light following different
patterns, alternating patterns, or just randomly. The more predictable the
pattern is, the easier for the player. The ability to communicate with other
watch towers and to detect footprints or noises are other behaviors that can
be enabled, disabled, or modified to change the difficulty of the game. In a
MOBA, the tower has the behavior to attack the enemies, by changing the target
priority, the game changes the difficulty drastically.

\subsection{Events} 

Events provide an alternative that doesn't required too many modifications to
the existing game codebase, as does the change in attributes and behaviors, but
their implementation
requires more design work than the previous two, and are easier to be spotted by
the
player if the implementation is poor. Events are  predefined occurrences
that arise under certain circumstances. For example, if the player is low on
health, and his performance is low, the next enemy will always drop a health potion.
Conversely, if the player's health is full, and his proficiency level is high,
the next
enemy hit will always deal critical damage.

\section{Models}\label{sec:mod}

In this section, we will introduce a small selection of existing
approaches for DDA.

\subsection{Metrics}\label{sec:met}

As mentioned before, it's important to identify the factors that directly influence the
player performance.
In the simpler use of metrics, a multiplier is applied to the variables that controls said
factors. The initial value for the multipliers is to be defined by the game developer, it’s
recommended to use a neutral multiplicative at the start of the game as there shouldn’t 
be any change on the factors yet.
For each relationship between the variables used to evaluate player performance
($EV$) and
the attributes that affect the player performance ($FV$), a
weight is defined. Note that the maximum number of weights needed is of $EV \times FV$,
in case that each attribute affects every variable, which means
adding variables or attributes
would greatly increment the number of weights to define. If possible, each variables should
be affected by a few attributes, and not all of them in order to reduce the number of weights.

These weights are added to the multiplier in function of the difference between player
performance ($PP$) and game difficulty ($GD$). The evaluation can be performed through
the use of thresholds~\cite{sutoyo2015dynamic} or multiplying the weight with said difference
$\frac{PP}{GD}$.

\subsection{Probabilistic Methods}

Probabilistic methods focus on predicting events on the game through the use of probabilistic
calculations and using the probabilities in a challenge function~\cite{xue2017dynamic}.

The probabilistic calculations are used to get the expected value of factors that directly affects
 the player performance and act accordingly before the player faces the challenge.

As an example, in case the player is reaching a zone with 40\% of his health, the probabilistic
 calculation will get the expected value of the total damage the player is going to suffer. This value
 will be returned to the challenge function that is going to evaluate whether the challenge is too difficult
 or too easy for the player and act accordingly before the player enters the zone.

Experiments show that the probabilistic method is
effective in games organized in stages~\cite{segundo2016dynamic} or levels~\cite{xue2017dynamic}, as it
 allows the AI to know which calculations it has to make based on the player
 current location and direction
 or to precalculate the values of each stage.

\subsection{Dynamic Scripting}
 
Dynamic Scripting is an online machine learning technique focused in the
modification of behaviors of agents
 in the game. Dynamic Scripts~(DS) are built from a set of rulebases, one for
 each
 type of agent to be modified.
Each time an agent is created, the associated rulebase is used to create a new script that controls its 
behavior.
Each rule of the rulebase has an associated weight that determine the chances of selecting the rule.
 The weights are adjusted based on a fitness function that evaluates the
 system's performance~\cite{spronck2006adaptive}.

\begin{figure}[t] 
\includegraphics[width=\columnwidth]{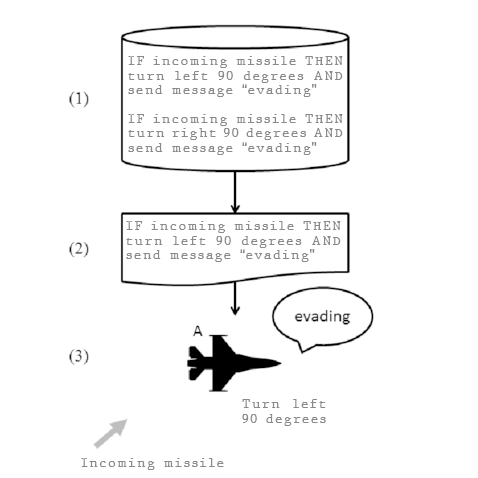}
\caption{Dynamic Scripting~\cite{Toubman2014Improving}}
\label{fig:DS}
 \end{figure}

Image~\ref{fig:DS} shows an example of a Dynamic Scripting system in action. The
character
 has an action rule-base. Variations on the same rules are performed to
 give the algorithm more
options. The DS selects rules from the rule base to create
the behavior script.

In order to
have a good performance, a Dynamic Scripting implementation has to meet
certain requirements~\cite{spronck2003online,spronck2006adaptive}:
\begin{description}
  \item[Speed:] Algorithms in online machine learning must be computationally
  fast since they
 take place during gameplay.
  \item[Effectiveness:] The created scripts should be at least as challenging as
  manually designed
 ones.
  \item[Robustness:] The learning mechanism must be able to cope with a
  significant
  amount of 
randomness inherent in most commercial gaming mechanisms.
  \item[Efficiency:] The learning process should rely on a small number of
  trials, since a player
 experiences a limited number of encounters with similar groups of opponents.
  \item[Clarity:]  Adaptive game AI must produce easily interpretable results,
  because game
 developers distrust learning techniques of which the results are hard to understand.
  \item[Variety:] Adaptive game AI must produce a variety of different
  behaviors, because 
agents that exhibit predictable behavior are less entertaining than agents that exhibit unpredictable behavior.
  \item[Consistency:]  The average number of learning opportunities needed for
  adaptive game
 AI to produce successful results should have a high consistency to ensure that their achievement is 
independent both from the behavior of the human player, and from random fluctuations in the learning process.
  \item[Scalability:] Adaptive game AI must be able to scale the difficulty
  level of its results to
 the skill level of the human player.
\end{description}

As Dynamic Scripting is a continually learning AI, it might keep improving until
it
reaches a point where it can always defeat
 the player. In DDA the objective isn’t to beat the
 player but to
 provide a fitting challenge, so in order to avoid
the AI surpassing the player skills some modifications have been added to
traditional Dynamic Scripting.

Weight Clipping is a technique where a maximum $W$ value is determined,
preventing the weights to grow over $W$~\cite{spronck2004enhancing}.
Normally, a set of rules with a large win rate will result in an AI with high
performance, thus increasing the weights 
of those rules and increasing the chances of being selected, resulting on an AI
that keeps defeating the player.
 By setting a low $W$ value, the chances of the rule set beaing selected are
 reduced, allowing the AI to pick tactics that make him
 lose against the player.

Top Culling technique, as Weight Clipping, sets a maximum $W$ value and allows the weights to grow limitless but
 weights with values that surpasses $W$ will not be selected to generate
 scripts~\cite{spronck2004enhancing}.

Another technique used to increase the player enjoyment is the Adrenaline Rush. This technique is based on the 
assumption that the player's learning rate is high at the start of the game but
drops through it. A learning limit is
 set to restrict the weights adjustment, and a maximum player fitness delta $P$
 is defined. By constantly measuring
 the player fitness and keeping track of the previous value a player fitness
 difference between the previous and current
 state can be calculated. When this player fitness delta drops below $P$ the learning limit of the weights is decreased so
 the AI doesn’t overgrow the player~\cite{Arulraj2010Adaptive}.

Finally, it’s possible for the Dynamic Scripting's fitness function to minimize
the difference between it's performance and the player's, rather than to
maximize absolute performance. This
way the AI
will adjust the weights to select rules that increase the chances of the game
 difficulty fitting the player skills, instead of the rules that make it more likely to win.

\section{Conclusion} 

Dynamic Difficulty Adjustment (DDA) is a technique that allows the developer to give the
 player a game that adapts itself to fit him, thus increasing player engagement. 
In recent years the amount of research in DDA has increased. As a result, there
are
 many approaches to implement it, with the main difference being the method used
 to
 change the attributes and behaviors of the game. Because of that, it is possible to have 
a general structure of how to implement the DDA.
The proposed structure is separated into two main parts, measuring the player proficiency 
and adjusting the game accordingly, using the method preferred by the developer.

Implementing a Dynamic Difficulty Adjustment system is a powerful tool that
allows for a better game experience. Its
implementation is based mainly in data collection from test subjects and the
player playing the game, and use of metric and mathematical functions to define
the changes to be performed. By adjusting variables and behaviors within the
game create challenges fit to the player, but careful planning is needed. Poor
implementation can result in having the opposite effect and overuse can cause
high processing consumption from the game.

\subsection{Future Work}
DDA is still a new field of research and much can be done to increase
 its effect in player engagement. New methods for implementation are always
 needed but research on optimization of already existing ones or creation of tools 
that enable an easier DDA implementation are also interesting.

Research can be done to reduce the process required for the evaluation of the 
fitness function, in order to be able to add more variables and create even more
 precise DDA systems. As well as the creation of a method that enable a low cost
 adjustment of weights allowing the developer to manipulate more factors of the game.

Integration of Behaviors Trees and Finite-State Machines to the Dynamic
Scripting approach, 
where all possible three nodes or states and transitions are preprogrammed, and the
 used ones are selected by the DDA system is another promising research area.

Finally the creation of tools that allows easy DDA implementation for all game styles 
and that can be used in the most popular frameworks or engines for game
development, such as Unity or Unreal, is a field that is still untouched.

\bibliographystyle{IEEETran}
%\balance
\newpage
\bibliography{refs}

\end{document}